\documentclass{article}

\usepackage{microtype}
\usepackage{graphicx}
\usepackage{subfigure}
\usepackage{booktabs} % for professional tables

\usepackage{hyperref}

\usepackage{bm}
\usepackage{amssymb}
\usepackage{amsmath}
\usepackage{multirow}
\usepackage{bbm}

\usepackage{ulem}

\usepackage[accepted]{icml2022}

\icmltitlerunning{Published as a conference paper at ICML 2022}

\begin{document}

\twocolumn[
\icmltitle{An Intriguing Property of Geophysics Inversion}

\icmlsetsymbol{equal}{*}

\begin{icmlauthorlist}
\icmlauthor{Yinan Feng}{lanl}
\icmlauthor{Yinpeng Chen}{microsoft}
\icmlauthor{Shihang Feng}{lanl}
\icmlauthor{Peng Jin}{Penn State,lanl}
\icmlauthor{Zicheng Liu}{microsoft}
\icmlauthor{Youzuo Lin}{lanl}
\end{icmlauthorlist}

\icmlaffiliation{lanl}{Earth and Environmental Sciences Division, Los Alamos National Laboratory,USA}
\icmlaffiliation{microsoft}{Microsoft Research, USA}
\icmlaffiliation{Penn State}{College of Information Sciences and Technology, The Pennsylvania State University, USA}

\icmlcorrespondingauthor{Youzuo Lin}{ylin@lanl.gov}

\icmlkeywords{Machine Learning, ICML}

\vskip 0.3in
]



\printAffiliationsAndNotice{}  % leave blank if no need to mention equal contribution

\begin{abstract}

Inversion techniques are widely used to reconstruct subsurface physical properties (e.g., velocity, conductivity) from surface-based geophysical measurements (e.g., seismic, electric/magnetic (EM) data). The problems are governed by partial differential equations~(PDEs) like the wave or Maxwell's equations. Solving geophysical inversion problems is challenging due to the ill-posedness and high computational cost. To alleviate those issues, recent studies leverage deep neural networks to learn the inversion mappings from measurements to the property directly.
% geophysical measurements to the geophysical property directly.

In this paper, we show that such a mapping can be well modeled by a \textit{very shallow}~(but not wide) network with only five layers. This is achieved based on our new finding of an intriguing property: \textit{a near-linear relationship between the input and output, after applying integral transform in high dimensional space.} In particular, when dealing with the inversion from seismic data to subsurface velocity governed by a wave equation, the integral results of velocity with Gaussian kernels are linearly correlated to the integral of seismic data with sine kernels. Furthermore, this property can be easily turned into a light-weight encoder-decoder network for inversion. The encoder contains the integration of seismic data and the linear transformation without need for fine-tuning. The decoder only consists of a single transformer block to reverse the integral of velocity.

Experiments show that this interesting property holds for two geophysics inversion problems over four different datasets. Compared to much deeper InversionNet~\cite{wu2019inversionnet}, our method achieves comparable accuracy, but consumes significantly fewer parameters. %For instance, on Marmousi data, our model only needs 1/20 parameters of those used in InversionNet. 
\end{abstract}

\section{Introduction}
\label{Introduction}

\begin{figure}[h]
\vskip 0.2in
\begin{center}
\centerline{\includegraphics[width=\columnwidth]{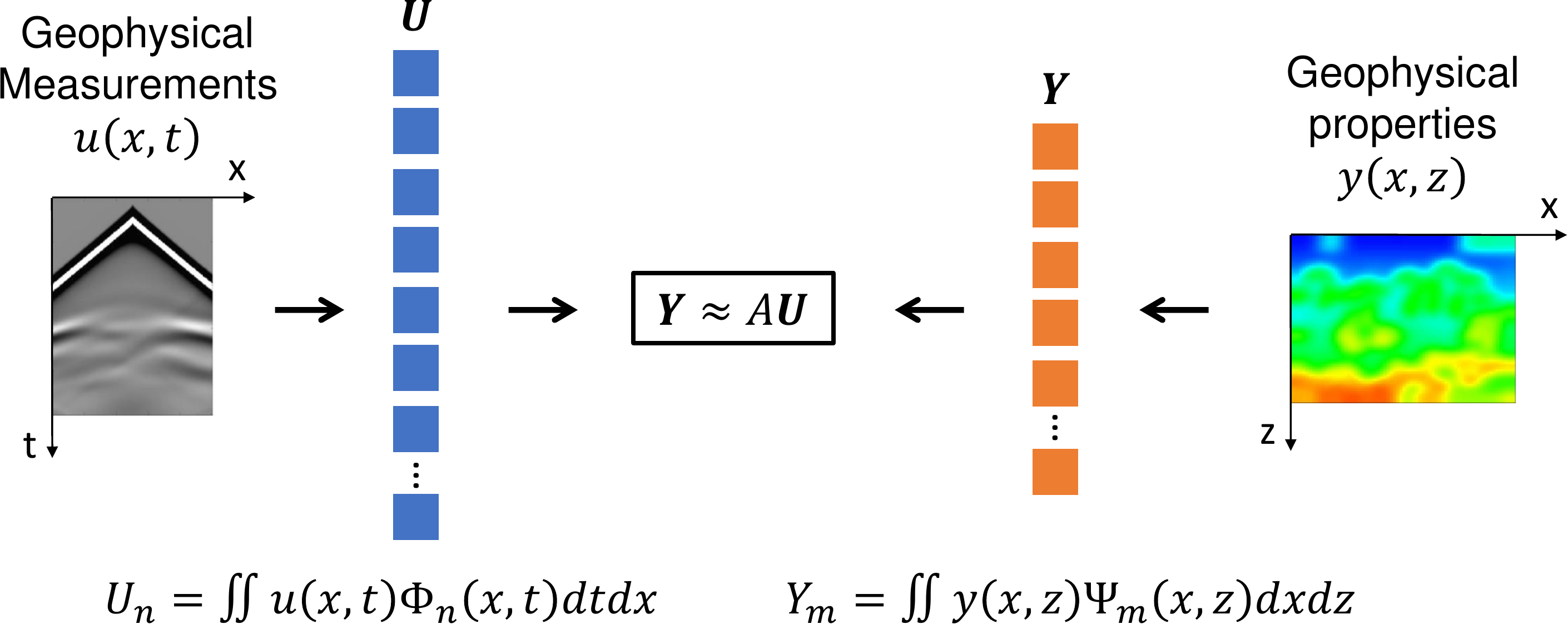}}
\caption{Illustration of the near-linear relation property between geophysical measurements and properties after applying integral transform. $\{\Phi_n\}$ and $\{\Psi_m\}$ are two families of kernels for integral transforms (e.g., sine and Gaussian). Here, the full waveform inversion from seismic data to velocity map is used as an example. 
%There is a near-linear relation between the frequency domain geophysical measurement and the embedding of geophysical properties after a Gaussian-kernel integral transform.
}
\label{teaser}
\end{center}
\vskip -0.2in
\end{figure}

Geophysics inversion techniques are commonly used to characterize site geology, stratigraphy, and rock quality. These techniques uncover subsurface layering and rock geomechanical properties~(such as velocity, conductivity), which are crucial in subsurface applications such as subsurface energy exploration, carbon capture and sequestration, groundwater contamination and remediation, and earthquake early warning systems.
Technically, these subsurface geophysical properties can be inferred from geophysical measurements~(such as seismic, electromagnetic~(EM)) acquired on the surface. Some underlying partial differential equations~(PDEs) between measurements and geophysical property exist, where inversion gets its name. For example, velocity is reconstructed from seismic data based on full waveform inversion~(FWI) of a wave equation, while conductivity is recovered from EM measurements based on EM inversion of Maxwell’s equations.% as shown in Figure \ref{teaser}.

However, these inversion problems can be rather challenging to solve, as they are ill-posed. Recent works study them from two perspectives: \textit{physics-driven} and \textit{data-driven}. The former approaches search for the optimal geophysical property~(e.g., velocity) from an initial guess, such that the generated geophysical simulations based on the forward modeling of the governing equation are closed to the real measurements~\cite{virieux2009overview,feng2019transmission+,feng2021multiscale}. These methods are computationally expensive as they require iterative optimization per sample. The latter methods~(i.e., data-driven approaches)~\cite{wu2019inversionnet}, inspired by the image-to-image translation task,  employ encoder-decoder convolution neural networks~(CNN) to learn the mapping between physical measurements and geophysical properties. \textit{Deep} network architecture that involves multiple convolution blocks is employed as both encoder and decode, which also results in heavy reliance on data and very high computational cost in training.

In this paper, we  found an intriguing property of geophysics inversion that can significantly simplify data-driven methods as:
\begingroup
\addtolength\leftmargini{-0.05in}
\begin{quote}
%\begin{center}
     \textit{Geophysical measurements~(e.g., seismic data) and geophysical property~(e.g., velocity map) have \textbf{near-linear} relationship in high dimensional space after \textbf{integral transform.}}
%\end{center}
\end{quote}
\endgroup

Let $u(x, t)$ denote a spatio-temporal geophysical measurement along horizontal $x$ and time $t$ dimensions, and $y(x, z)$ denote a 2D geophysical property along horizontal $x$ and depth $z$. Since, in practice, geophysical measurement is mostly collected at the surface, and people want to invert the subsurface geophysical property, measurement $u$ only contains spatial variable $x$, while property $y$ includes $(x,z)$.
As illustrated in Figure~\ref{teaser}, the proposed property can be mathematically represented as follows: 
\begin{align}
& \bm{U}=[U_1, \dots, U_N]^T, \;\; U_n = \iint u(x, t)\Phi_n(x, t)dxdt, \nonumber\\
& \bm{Y}=[Y_1, \dots, Y_M]^T, \;\; Y_m = \iint y(x,z)\Psi_m(x,z)dxdz, \nonumber \\
& \bm{Y} \approx \bm{A}\,\bm{U},
\label{eq:linear-property}
\end{align}
where $\Phi_n$ and $\Psi_m$ are kernels for integral transforms. After applying integral transforms, both geophysical measurement $u(x, t)$ and property $y(x, z)$ are projected into high dimensional space~(denoted as $\bm{U}$ and $\bm{Y}$), and they will have a near-linear relationship ($\bm{Y} \approx \bm{A}\,\bm{U}$). Note that the kernels ($\{\Phi_n\}, \{\Psi_m\}$) are \textit{not} learnable, but well-known analytical kernels like sine, Fourier, or Gaussian.

Interestingly, this intriguing \textit{property} can significantly simplify the encoder-decoder architecture in \textit{data-driven} methods. The encoder only contains the integral with kernel $\{\Phi_n\}$ followed by a linear layer with weight matrix $\bm{A}$ in Eq. \ref{eq:linear-property}. The decoder just uses a single transformer \cite{NIPS2017_transformer} block followed by a linear projection to reverse the integral with kernels $\{\Psi_m\}$. This results in a much shallower architecture. In addition, the encoder and decoder are learnt separately. The matrix $\bm{A}$ in encoder can be directly solved by pseudo inverse and is frozen afterward. Only the transformer block and following linear layer in the decoder are learnt via SGD based optimizer.

Our method, named InvLINT (Inversion via LINear relationship between INTegrals), achieves comparable (or even better) performance on two geophysics inversion problems (seismic full waveform inversion and electric/magnetic inversion) over four datasets (Kimberlina Leakage \cite{jordan2017characterizing}, Marmousi \cite{feng2021multiscale}, Salt \cite{yang2019deep}, and Kimberlina-Reservoir \cite{Development-2021-Alumbaugh}), but uses significantly less parameters than prior works.
For instance, on Marmousi, our model only needs 1/20 parameters, compared to previous InversionNet. 

\section{Background}
The governing equation of the seismic full waveform inversion is acoustic wave equation \cite{schuster2017seismic},
\begin{equation}
 \nabla^2p(\bm{r},t) -\frac{1}{c^2(\bm{r})} \frac{\partial^2}{\partial t^2}p(\bm{r},t) = s(\bm{r},t), \label{Acoustic}
\end{equation}
where $\bm{r} = (x,z)$ represents the spatial location in Cartesian coordinates ($x$ is the horizontal direction and $z$ is the depth), $t$ denotes time, $c(\bm{r})$ is the velocity map, $p(\bm{r},t)$ represents the pressure wavefield, $\nabla^2$ is the Laplacian operator, and $s(\bm{r}, t)$ is the source term that specifies the location and time history of the source.
% Full Waveform Inversion (FWI) is a methodology that calculates high-resolution the subsurface velocity maps $c(\bm{r})$ from the raw recorded seismic data $p(\bm{\tilde{r}}, t)$, where $\bm{\tilde{r}}$ represents the locations of the seismic receivers. 

For the EM forward modeling, the governing equation is the Maxwell's Equations \cite{commer2008new},
\begin{eqnarray}
\sigma\mathbf{E}-\nabla\times\mathbf{H}&=&-\mathbf{J}, \\\nonumber
\nabla\times\mathbf{E}+i\omega\mu_{0}\mathbf{H}&=&-\mathbf{M},
\label{eq:EMForward}
\end{eqnarray}
where $\mathbf{E}$ and $\mathbf{H}$ are the electric and  magnetic fields. $\mathbf{J}$ and $\mathbf{M}$ are the electric and magnetic sources. $\sigma$ is the electrical conductivity and $\mu_{0}=4\pi\times10^{-7} \Omega\cdot s/m$ is the magnetic permeability of free space.

\section{Methodology}
In this section, we use seismic full waveform inversion~(from seismic data to velocity) as an example to illustrate our derivation of the linear property after integral transforms. We will also show the encoder-decoder architecture based on this linear property. Empirically, our solution is also applicable to EM inversion~(from EM data to conductivity).

\subsection{Near-Linear Relationship between Integral Transformations}
In the following part, we will show the seismic data and velocity maps
have the near-linear relation  after integral transformation like the format of Equation \ref{eq:linear-property}. The seismic data $p$ and velocity map
$c$ are governed by the wave equation~(Equation \ref{Acoustic}).
Note that seismic data $p$ and velocity map
$c$ in wave equation corresponds to the input $u$ and output $y$ in Equation \ref{eq:linear-property}, respectively.
 %In practice, one can hardly get the full field measurement. Thus, we consider the two-dimensional case.

\textbf{Rewriting wave equation in Fourier series:} Similar to constant coefficients PDEs, we assume spatial variable $\bm{r} = (x,z)$ and temporal variable $t$ are separable, i.e., $p(x,z,t) = p_1(x,z)p_2(t)$, and $s(x,z,t)=s_1(x,z)s_2(t)$. Thus, Equation~\ref{Acoustic} is rewritten as
\begin{align}
 c^2(x,z)(\nabla^2p_1(x,z)p_2(t) - s_1(x,z)s_2(t)) \nonumber\\
 = \frac{\partial^2}{\partial t^2}(p_1(x,z)p_2(t)). \label{math1}
\end{align}

Next the temporal parts $p_2(t)$ and $s_2(t)$ are represented as Fourier series: $p_2(t)=\sum_{n=1}^N B_n e^{j2\pi nt}$ and $s_2(t)=\sum_{n=1}^N G_n e^{j2\pi nt}$. This turns Equation~\ref{math1} as:
\begin{align}
 \sum_{n=1}^N c^2(x,z)(\nabla^2p_1(x,z)B_n - s_1(x,z)G_n)e^{j2\pi nt} \nonumber\\
 = \sum_{n=1}^N 4\pi^2n^2p_1(x,z)B_ne^{j2\pi nt}. 
 \label{5}
\end{align}
To make sure both sides have the same coefficient for each $n$, the aggregation $\sum_{n=1}^N$ and $e^{j2\pi nt}$ can be removed from Equation \ref{5} as:
\begin{align}
 c^2(x,z)(\nabla^2p_1(x,z)B_n - s_1(x,z)G_n) \nonumber\\
 = 4\pi^2n^2p_1(x,z)B_n. 
\end{align}

By further integrating over $x$, we have 
% \begin{align}
%   c^2(x,z)(\nabla^2p_1(x,z)B_n - s_1(x,z)G_n) = \nonumber\\
%   4\pi^2n^2p_1(x,z)B_n.
% \end{align}
% Then, we integrate $x$ on both sides of the equation, and take $z=0$ (we will discuss this later). Moreover, since $B_n$ and $G_n$ are complex number, we take module on both side. 
% After arranging the equation, we get \bigg|_{z=0}
\begin{align}
  &\frac{1}{4\pi^2n^2}\int c^2(x,z) \left \lvert \nabla^2p_1(x,z)B_n - s_1(x,z)G_n  \right \rvert dx, \nonumber\\
  &= \int p_1(x,z) \left\lvert B_n  \right\rvert dx, \nonumber \\
  &= \left\arrowvert\iint \underbrace{p_1(x,z)p_2(t)}_{Seismic \,  data} \underbrace{e^{-j2\pi nt} \vphantom{p_1(x,z)p_2(t)}}_{Fourier \,  kernel}dt  dx \right\arrowvert,\label{math2}
\end{align}
where $|\cdot|$ is the modulus operator of complex numbers and $B_n = \int p_2(t)e^{-j2\pi nt}dt$ are the Fourier coefficients. Note that since $B_n$ and $G_n$ are complex numbers, we take module on both sides. Here, taking the real or imaginary part, rather than modulo, does not affect our conclusions.
Now, the right hand of Equation~\ref{math2} is in the same format with $U_n$ in Equation \ref{eq:linear-property}. The kernel function $\Phi_n(x,t) = e^{-j2\pi nt}\mathbbm{1}(x)$, where $\mathbbm{1}(x)=1$ for all $x$.

% To build the near-linear relationship, for the next step, we need to simplify the complicated kernel function in the left hand side of Equation~\ref{math2} by approximation.
% a near-linear relation between $\int p_1(x,z) \lvert B_n \rvert dx$ and $c^2(x,z)$ after a integral transformation.
%$p_i(x,z)$'s Fourier coefficient $B_n$

%\subsubsection{Single kernel approximation:}

\textbf{Approximation by Integral over} $z$:
In reality the seismic data $p(x,z,t)$ is mostly collected at the surface~($z=0$). Thus, the right-hand side of Equation~\ref{math2} is computable at $z=0$.
However, the left-hand side is hard to calculate, since $\nabla^2p_1(x,z)$ and $s_1(x,z)$ are unknown. Here, we hypothesize that the left-hand side at $z=0$ can be approximated by leveraging velocity map at multiple depth positions as:
%Moreover, since we want to invert not only the surface but also subsurface geophysical properties, we make the following hypothesis:
% we first approximate the complex integration result in Equation~\ref{math2} at $z=0$ by an integration over $x$ and $z$ axis. 

%\textbf{Hypothesis}: The integral transformation of $c^2(x,z)$ along $x$ axis (left-hand side of Equation \ref{math2}) at $z=0$ can be approximated by an integral transformation of $c^2(x,z)$ along $x$ and $z$ axis with a kernel $F_n(x,z)$. Formally,
\begin{align}
  \frac{1}{4\pi^2n^2}\int c^2(x,z) \left \lvert \nabla^2p_1(x,z)B_n - s_1(x,z)G_n  \right \rvert dx\bigg|_{z=0} \nonumber\\
  \approx \iint c^2(x,z)F_n(x,z)dxdz, \label{math3}
\end{align}
where $F_n(x,z)$ is the kernel function.

This hypothesis (Eq. 7--8) bridges integral transforms of the seismic data ($\iint p(t,x,z) e^{-j2\pi nt} dt  dx |_{z=0}$) and velocity maps ($ \iint c^2(x,z)F_n(x,z)dxdz$) via an auxiliary function  $\frac{1}{4\pi^2n^2}\int c^2(x,z) \left \lvert \nabla^2p_1(x,z)B_n - s_1(x,z)G_n  \right \rvert dx |_{z=0}$.

It has two parts: (a) the double integral of velocity maps equals the auxiliary function, and (b) the 2D kernel $F_n(x,z)$ can be estimated by a set of basis functions, so we can further calculate the inverse problem we want to solve. The existence of $F_n(x,z)$ to achieve the former equality can be validated by a special case $F_n(x,z)=\frac{1}{4\pi^2n^2}\left \lvert \nabla^2p_1(x,z)B_n - s_1(x,z)G_n  \right \rvert \delta(z)$
 where $\delta(z)$ is an impulse function. 
 The latter may weaken the former assertion of equality, but the misfit is likely small, as velocity map is continuous at most $(x, z)$ positions and seismic data $p_1(x, z)$ and source $s_1(x, z)$ in the auxiliary function has strong correlation along $x$ and $z$. Our experimental results over three datasets
 empirically validate this hypothesis.

% Here, The basic idea is that velocity map $c(x,z)$, seismic data $p(x,z,t)$ and source $s(x,z,t)$ are highly correlated. Thus, we can use the information for $z>0$ to infer the transformation at $z=0$. 
% for the real geophysical measurements and properties, $x$ and $z$ axis 

\textbf{Further simplification by a single kernel family:} 
As discussed above, we simplify $F_n(x,z)$ as a weighted sum of a series of basis functions:
\begin{align}
  F_n(x,z) = \sum_{m=1}^M d_{n,m}\Psi_m(x,z), \label{math4}
\end{align}
where $d_{n,m}$ is the weight and $\Psi_m(x,z)$ is the basis function.
By further plugging Equations \ref{math3} and \ref{math4} into Equation \ref{math2}, we get
\begin{align}
  &\sum_{m=1}^M d_{n,m}\iint c^2(x,z) \Psi_m(x,z)dxdz \nonumber\\
  &\approx \left\lvert\iint p(x,z,t)e^{-j2\pi nt}dt  dx \right\rvert _{z=0}.
  \label{math5}
\end{align}
%Note that after the above derivation, the left-hand side of the Equation \ref{math5} has become a weighted sum of transformed $c^2(x,z)$. %and we have successfully built a near-linear relationship between seismic data and velocity map in a high dimensional space after integral transforms.
%which is what we would expect as in Equation~\ref{eq:linear-property}. W

\textbf{Relation to Equation 1:}
Equation~\ref{math5} is special case of Equation~\ref{eq:linear-property}
% Now, we have successfully used wave equation as a special case of Equation~\ref{eq:linear-property} to demonstrate that there exists a \textit{near-linear relation} between the geophysical measurement~(i.e., seismic data) and geophysical property~(i.e., velocity map) after mapping them to a high dimension space with two integral transformations.
% The raw recorded measurement data $p(\bm{\tilde{r}}, t)$ is often a 2-D spatio-temporal data collected at the surface~($z=0$), where $p(\bm{\tilde{r}}, t) = p(x, t)$~($\bm{\tilde{r}}$ represents the locations of the seismic receivers). 
We can, therefore, express Equation \ref{math5} in the form of  Equation \ref{eq:linear-property} by letting:
\begin{align}
  y(x,z) = c^2(x,z), \qquad &u(x,t) = p(x,t), \nonumber \\
  A = D^{\dagger}, \qquad &\Phi_n(x,t) = e^{-j2\pi nt}\mathbbm{1}(x), \nonumber
\end{align}
where $A$ is pseudo inverse of matrix $D$ and $D = \left [d_{n,m} \right ]_{N\times M}$ is the matrix format. %Note that we replace Fourier transform with cosine/sine kernel~(discrete cosine/sin transform) for ease of the implementation. 

In particular, $\bm{U}=[U_1, \dots, U_M]^T$ and $\bm{Y}=[Y_1, \dots, Y_N]^T$ are the high dimensional embeddings of the measurement and geophysical property. 
$\{\Phi_n\}$ is chosen as cosine/sine or Fourier transform; while, based on the experiments, Gaussian kernel becomes our choice of the $\{\Psi_m\}$ to embed the spatial information in the geophysics property. It is true that the hypothesis may seem strong, however, its validity can be supported via our extensive experimental results using multiple datasets and various PDEs.

\subsection{Simplified Encoder-Decoder Architecture}
\begin{figure}[t]
\vskip 0.2in
\begin{center}
\centerline{\includegraphics[width=\columnwidth]{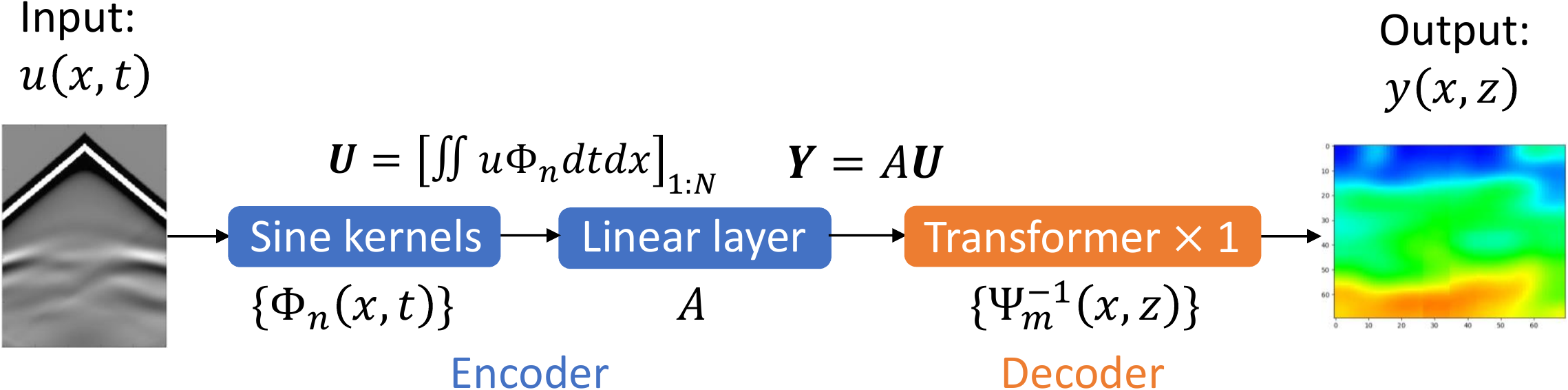}}
\caption{Schematic illustration of our proposed method, using seismic FWI as an example. The linear regression for two transformed embeddings is solved by pseudo inverse and is frozen afterward. The decoder is trained via SGD-based optimizer.}
\label{workflow}
\end{center}
\vskip -0.2in
\end{figure}

Based on the proposed mathematical property as shown in Equation~\ref{eq:linear-property}, we can easily design a simple network architecture, accordingly. The encoder plays exactly the same role as the right-hand side of Equation~\ref{eq:linear-property}, while the decoder, with a neural network, approximates the inverse mapping of the integral transformation ($\Psi_m^{-1}(x,z)$). The structure~(Figure~\ref{workflow}) of our InvLINT is described below.

\textbf{Encoder:}
As illustrated in Figure~\ref{workflow}, we design the encoder exactly the same to Equation~\ref{eq:linear-property}, where an integral transformation with kernel $\{\Phi_n\}, n\in[1,N]$ is first implemented and followed by a linear layer represented by $A$. With such a simple linear relation, one can easily map the input measurement to the embedding of the output.

\textbf{Decoder:}
There are many kernel functions~(like Gaussian kernel), which does not have a close form inverse transformations. Instead, we use a shallow decoder network to approximate such a pseudo-inverse. To achieve this, we first use a linear layer $L_1$ to map $Y$ to a more compact embedding and tile it a grid with the shape $R^{h\times w \times k}$. Here, $h$ and $w$ are the size of the velocity map with $32$ times downsampling, and $k$ is the number of channels. After that, $L_1(Y)$ is input into a 1-layer transformer, with patch size of $1\times1\times k$. This shallow transformer is the only nonlinear part of our decoder.

The last parts of the model are a linear layer $L_{r}$. It upsamples each $1\times1\times k$ patch to a $(32+d)\times(32+d)$ block\footnote{Since the size of output may not be divided exactly by $32$, the recovered shape will be slightly different for different datasets.}, where $d$ is a small integer. The final predicted velocity map $\hat{c}$ can be construed by stitching all $h\times w$ blocks together. The purpose of this is to recover the output to the original shape with overlaps among blocks to remove the block effect.

\subsection{Training}
Because of the near-linear relation, we can easily solve the linear layer in the encoder, $L_1$, with the least squares method. Specifically, we first compute the embedding of both encoder and decoder by integral transformations, calculate the solution of matrix $A$, and freeze it while training the decoder. The decoder is trained by an SGD-based optimizer. The loss function of our InvLINT is a pixel-wise MAE loss given as
\begin{eqnarray}
\mathcal{L}(\hat{c}, c) = \ell_1(\hat{c}, c).
\end{eqnarray}
Peng et al. \cite{jin2021unsupervised} find that combining MAE, MSE, and perceptual loss together is helpful to improve the performance. However, to make a fair comparison with the previous work, we only use MAE as our loss function.

\section{Experiment}
In this section, we present experimental results of our proposed InvLINT evaluated on four datasets and compare our method with the previous works, InversionNet~\cite{wu2019inversionnet} and VelocityGAN~\citet{zhang2019velocitygan}. We also discuss different factors that affect the performance of our method.
%\begin{tabular}[c]{@{}l@{}}InversionNet \\ \cite{wu2019inversionnet}\end{tabular}
\begin{table*}[!ht]
\centering
\setlength{\tabcolsep}{1.8mm}
\begin{tabular}{l|l|l|l|l|l|l}
\hline
{\textbf{Dataset}} & {\textbf{Model}} & {\textbf{MAE$\downarrow$}} & {\textbf{MSE$\downarrow$}} & {\textbf{SSIM$\uparrow$}} & {\textbf{\#Parameters}} & {\textbf{FLOPs}} \\ \hline
\multirow{3}{*}{\begin{tabular}[c]{@{}l@{}}Kimberlina \\ Leakage\end{tabular}} & InversionNet \cite{wu2019inversionnet} & 9.43              & 1086.99  & \textbf{0.9868} & 15.81M                   & 563.52M                  \\ \cline{2-7}
                                    & VelocityGAN \citet{zhang2019velocitygan} & 9.73              & \textbf{1026.27}  & 0.9863 & 16.99M                   &   1.31G                \\ \cline{2-7}
                                    & InvLINT (Ours)         & \textbf{8.13}     & 1534.60           & 0.9812          & \textbf{1.49M (9.4\%)}  & \textbf{44.30M (7.9\%)} \\ \hline
\multirow{3}{*}{Marmousi}           & InversionNet \cite{wu2019inversionnet} & 149.67            & 45936.23          & 0.7889          & 24.41M                   & 189.58M                  \\ \cline{2-7}
                                    & VelocityGAN \citet{zhang2019velocitygan} & \textbf{124.62}              & \textbf{30644.31}  & \textbf{0.8642} & 25.59M                   & 259.49M                  \\ \cline{2-7}
                                    & InvLINT (Ours)         & 136.67   & 36003.43 & 0.7972 & \textbf{1.45M (5.9\%)}  & \textbf{9.31M (4.9\%)}  \\ \hline
\multirow{3}{*}{Salt}               & InversionNet \cite{wu2019inversionnet} & 25.98    & \textbf{8669.98}  & \textbf{0.9764} & 13.74M                   & 32.37M                   \\ \cline{2-7}
                                    & VelocityGAN \citet{zhang2019velocitygan} &  332.62      & 145669.11  & 0.7760 & 14.92M                   &   65.98M               \\ \cline{2-7}
                                    & InvLINT (Ours)         & \textbf{24.60}             & 8840.79           & 0.9742          & \textbf{1.62M (11.8\%)} & \textbf{5.98M (18.5\%)} \\ \hline
\multirow{3}{*}{\begin{tabular}[c]{@{}l@{}}Kimberlina- \\ Reservoir\end{tabular}} & InversionNet \cite{wu2019inversionnet} & 0.01330           & 0.000855          & 0.9175          & 0.30M                    & 1.20G                    \\ \cline{2-7}
                                    & VelocityGAN \citet{zhang2019velocitygan} & 0.01313              & 0.000688  & 0.8611 & 1.48M                   & 3.95G                  \\ \cline{2-7}
                                    & InvLINT (Ours)         & \textbf{0.00703} & \textbf{0.000537} & \textbf{0.9370} & \textbf{0.16M (53.3\%)} & \textbf{96.10M (8.0\%)} \\ \hline 
\end{tabular}
\caption{Quantitative results evaluated on four datasets in terms of MAE, MSE and SSIM, the number of parameters and FLOPs. The percentages indicate the ratio of \#Parameters (FLOPs) required by InvLINT to that required by InversionNet. Our
InvLINT achieves comparable (or even better) inversion accuracy comparing to the InversionNet and VelocityGAN with a much smaller number of parameters and FLOPs.}
\label{main-result}
\end{table*}

\subsection{Implementation Details}
\subsubsection{Datasets}
In experiments, we verify our method on four datasets, of which three of them are used for seismic FWI, and one of which is for an EM inversion.

\textbf{Kimberlina-Leakage:} The geophysical properties were developed under DOE’s National Risk Assessment Program~(NRAP). It contains 991 CO$_2$ leakage scenarios, each simulated over a duration of 200 years, with 20 leakage velocity maps provided~(i.e., at every ten years) for each scenario \cite{jordan2017characterizing}.
Excluding some missing velocity maps, the data are split as 807/166 scenarios for training and testing, respectively.
The size of the velocity maps is $401\times14$1 grid points, and the grid size is 10 meters in both directions. To synthesize the seismic data, nine sources are evenly distributed along the top of the model, with depths of 5 $m$. The seismic traces are recorded by 101 receivers positioned at each grid with an interval of 15 $m$. The source frequency is 10 Hz. Each receiver collects 1251-timestep data for 1 second.

\textbf{Marmousi:} We apply the generating method in \citet{jin2021unsupervised}, which follows \citet{feng2021multiscale} and adopts the Marmousi velocity map as the style image to construct this low-resolution dataset. This dataset contains 30K with paired seismic data and velocity map. 24k samples are set as the training set, 3k samples are used as the validation set, and the rest are the testing set. The size of the velocity map is $70\times70$, with the 10-meter grid size in both directions. The velocity ranges from $1,500 m/s$ to $4,700 m/s$. There are $S=5$ sources placed evenly with a spacing of 170 $m$. The source frequency is 20 Hz. The seismic data are recorded by 70 receivers with a receiver interval of 10 $m$. Each receiver collects 1,000-timestep data for 1 second. 

\textbf{Salt:} The dataset contains 140 velocity maps \cite{yang2019deep}. We downsample it to $40\times60$ with a grid size of 10 $m$, and the splitting strategy 120/10/10 is applied. The velocity ranges from $1,500 m/s$ to $4,500 m/s$. There are also $S=5$ sources used, with 12-Hz source frequency and a spacing of 150 $m$. The seismic data are recorded by 60 receivers with an interval of 10 $m$, too. Each receiver collects 600-timestep data for 1 second. 

\textbf{Kimberlina-Reservoir:} The geophysical properties were also developed under DOE’s NRAP. It is based on a potential $\mathrm{CO_{2}}$ storage site in the Southern San Joaquin Basin of California \cite{Development-2021-Alumbaugh}. We use this dataset to test our method in the EM inversion problem. In this data, there are 780 EM data as the geophysical measurement, and corresponding conductivity as the geophysical property. We use 750/30 as training and testing.
EM data are simulated by finite-difference method \cite{commer2008new} with two sources location at $x=2.5$ $km$, $z=3.025$ $km$ and $x=4.5$ $km$, $z=2.5$ $km$. There are $S=8$ source frequencies from $0.1$ to $8.0$ Hz and recorded with its real and imaginary part. The conductivity is with the size of $351\times601$ ($H\times{W}$), where the grid is 10 $m$ in all dimensions.
%In order to distinguish it from Kimberlina-Leakage dataset, we will refer to it as \textit{EM dataset} later.

\subsubsection{Training Details}
The input seismic data and EM data are normalized to the range [-1, 1].
In practice, to supply more information, it always uses multiple sources to measure, where $s \in [1, \cdots S]$ is the index of different sources. After integration, all sources vectors will be concatenated. For the seismic data, we use $\Phi_n(x,t) = sin(n\pi t)\mathbbm{1}(x)/(x_{max}-x_{min})$ as the kernel function. However, for the EM data, since the raw data are already in the frequency domain and the input size is small, we skip the integral transformation step. 
%, where $n_S = nS+1$ and $n \in \{0,1,2, \cdots, \lceil\frac{2048}{S}\rceil\}$ Since, for different datasets, the number of source is different, the number of dimensions is slightly different.

The Gaussian kernel can be represented as $\Psi_m(x,z) = \exp{\frac{-\|(x,z)-\mathbf{\mu_m}\|_2^2}{2\sigma^2}}$. We let $\mathbf{\mu_m}$ distribute evenly over the output shape. Then, the $\sigma$ is set equal to the distance of adjacent $\mathbf{\mu}$. 
When applying Ridge regression to solve the linear layer in the encoder, and set the regularization parameter $\alpha=1$.
%set $M=512$ (except, $32$ for EM data), and

%\yinan{Empirically, for FWI, we find using the integral transformation of velocity map $c(x,z)$ yields similar results to its square. Thus, for the simplicity of the model, we will use $c(x,z)$ in all experiments.}

We employ AdamW \cite{loshchilov2018decoupled} optimizer with momentum parameters $\beta_1 = 0.5$, $\beta_2 = 0.999$
and a weight decay of $1 \times 10^{-4}$ to update decoder parameters of the network. The initial learning rate is
set to be $1 \times 10^{-3}$, and we decay the learning rate with a cosine annealing \cite{loshchilov2016sgdr}, where $T_0=5$, $T_{mult}=2$ and the minimum learning rate is set to be $1 \times 10^{-3}$. The size of every mini-batch is set to be 128. We implement our models in Pytorch and train them on 1 NVIDIA Tesla V100 GPU.

\subsubsection{Evaluation Metrics}
We apply three metrics to evaluate the geophysical properties generated by our method: MAE, MSE and Structural Similarity (SSIM). In the existing literature \cite{wu2019inversionnet, zhang2020data}, both MAE and MSE have been employed to measure the pixel-wise error. SSIM is also considered to measure the perceptual similarity~\cite{jin2021unsupervised}, since both velocity maps and conductivity have highly structured information, and degradation or distortion can be easily perceived by a human. Note that when calculating MAE and MSE, we denormalize geophysical properties to their original scale while we keep them in the normalized scale $\left[-1, 1\right]$ for SSIM according to the algorithm.

Moreover, we also employ two common metrics to measure the complexity and computational cost of the model: the number of parameters (\#Parameters) and Floating-point operations per second~(FLOPs).

\subsection{Main Results}
Table~\ref{main-result} shows the comparison results of our method with InversionNet on different datasets. Overall, our method achieves comparable or even better performance with a smaller amount of parameters and lower FLOPs. Below, we will provide in detail the comparison of all four datasets. It may be worthwhile mentioning that FWI is a quantitative inversion technique, meaning that it will yield both the shape and the quantitative values of the subsurface property. 

\begin{figure}[!ht]
\begin{center}
\centerline{\includegraphics[width=\columnwidth]{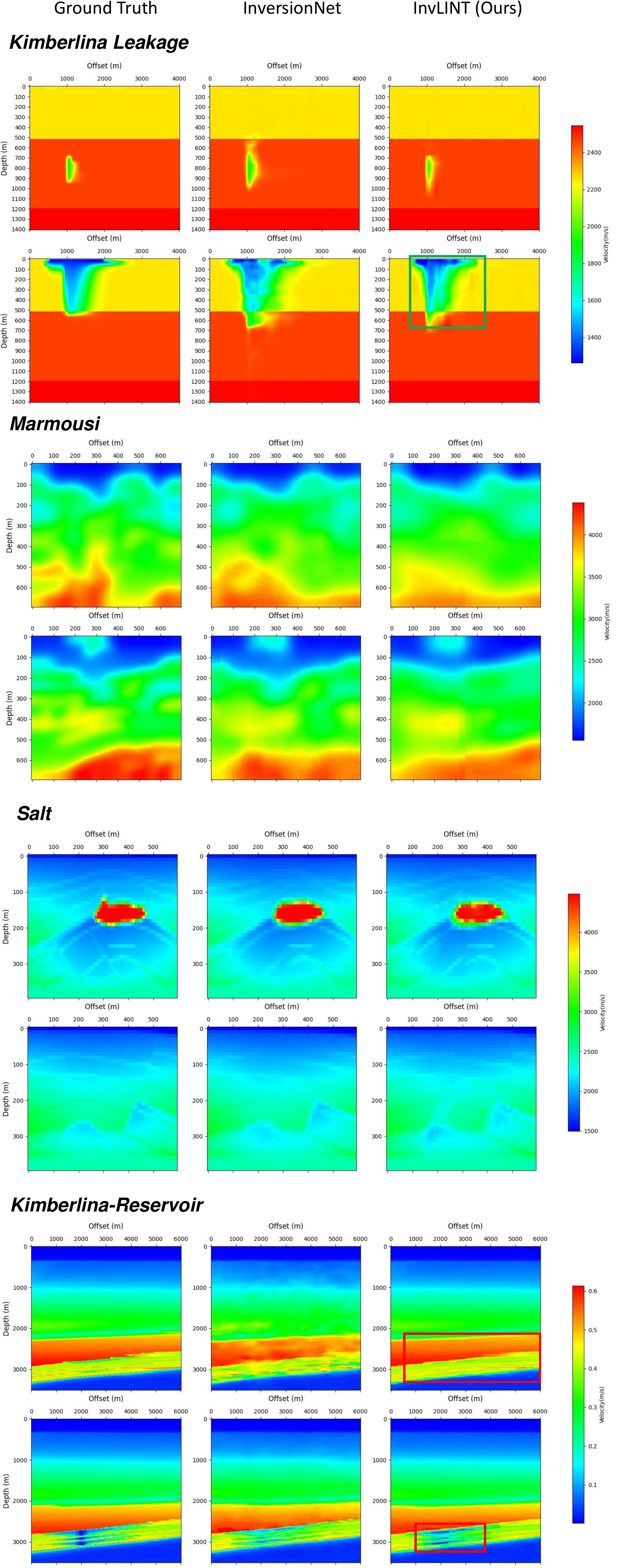}}
\caption{Illustration of results evaluated on four datasets} %(a) Results on Kimerlina CO$_2$; (b) Results on Marmousi; (c) Results on Salt; (d) Results on Kimberlina-Reservoir. Our InvLINT achieves good results on all four datasets.
\label{main-fig}
\end{center}
\vskip -0.2in
\end{figure}

\textbf{Results on Kimerlina-Leakage:} 
Compared to InversionNet and VelocityGAN, our method outperforms in MAE, slightly worse in MSE and SSIM. However, our InvLINT only needs less than $1/10$ parameters and FLOPs. This demonstrates the power of our model, and further validates the properties we found. The velocity maps inverted by ours and InversionNet are shown in the first two rows of Figure~\ref{main-fig}. In the second example, despite of some noise produced by our method in the background, the CO$_2$ leakage plume~(most important region as boxed out in green) has been very well imaged. Compared to ground truth, our method yields even better quantitative values than that obtained by InversionNet. 

% The second example shows our method may produce slight noise in the background. However, it does successfully invert the CO$_2$ leakage plume~(as boxed out in green) in both shape and value, while the inversion accuracy is better than the InversionNet as highlighted by the green square.

\textbf{Results on Marmousi:} 
Marmousi is a more challenging dataset due to its more complex structure. Compared to InversionNet, our method outperforms in all three metrics with significantly less computational and memory cost~(about $1/20$ parameters and FLOPs). This result again demonstrates not only the power of our model but also the validity of the near-linear relationship that we found. However, in such a large and complex dataset, VelocityGAN outperform others, where the GAN structure helps generating better results. The velocity maps inverted by ours and InversionNet are illustrated in the third and fourth rows of Figure~\ref{main-fig}. Our InvLINT and InversionNet perform comparably in both the shallow and deep regions compared to the ground truth. 

\textbf{Results on Salt:} 
Compared to InversionNet, our method outperforms in MAE, and is slightly worse in MSE and SSIM with a very small gap. Moreover, our method uses $1/8$ parameters and $1/5$ FLOPs to those of InversionNet. Note that, in this challenging dataset, which only has a small number of samples, VelocityGAN cannot converge well and yields bad results. This is a side effect of its complex structure. The velocity maps inverted by ours and InversionNet are illustrated in the fifth and sixth rows of Figure~\ref{main-fig}. Consistent with quantitative results, both methods generate similar results. In the shallow region, our method output a slightly clear structure; but in a deeper region~(e.g., the red region in the first example), the output of InversionNet is a little close to the ground truth. However, the overall difference can be hard to distinguish. Our method achieves comparable results with much less complexity.

\textbf{Results on Kimberlina-Reservoir:} 
Compared to InversionNet and VelocityGAN, our method outperforms in all three metrics, with $1/2$ parameters and $1/12$ FLOPs to those of InversionNet. Because of the compact input, all model utilize the much smaller number of parameters. However, due to the simple architecture, InvLINT yields significantly fewer FLOPs but achieves better inversion accuracy.
The conductivity results inverted by different models are shown in the last two rows of Figure~\ref{main-fig}. Contrary to previous results on the Kimberlina-Leakage dataset, our model yields clearer results. In the first example, we can see that the outputs of our model are less noisy; and in the second case, InvLINT inverts the deep region more precisely, as highlighted by the red squares. This is also consistent with the quantitative results.

At the same time, we find that the number of parameters of our model varies less for the same inverse problem. The number of model parameters is relatively independent of data size. In contrast, the previous methods are greatly affected by the input and output sizes. 
Moreover, our model not only requires fewer parameters, but also enables more \textit{efficient} training and inference. When training on Marmousi dataset using 1 GPU (NVIDIA Quadro RTX 8000), our model is \textbf{\textit{9 times faster}} than InversionNet/VelocityGAN (1 hour vs. 9 hours). We also tested inference runtime with batch size 1 on a single thread of an Intel(R) Xeon(R) CPU Gold 6248 v3 (2.5GHz). Our model is \textbf{\textit{16 times faster}} than InversionNet/VelocityGAN (5 ms vs. 80ms). The small model size is suitable for memory-limited mobile devices. More visualization results are provided in the Appendix for readers who might be interested.

\subsection{Ablation Tests}
In this part, we will test how different kernel functions and network architectures will influence the performance of our method. We put our default setting at the first row of each table. For ease of illustration, we only provide results on Marmousi. Results on Kimberlina Leakage are given in the Appendix. %Moreover, we will evaluate our model's robustness with Gaussian Noise or missing traces.

\textbf{Different Encoder Kernels}

First, we conduct experiments by replacing the 1D sine kernel in the encoder to different 2D sine kernels. The quantitative results are shown in the Table~\ref{enkernel}. By comparing the results in Marmousi and the results in Kinberlina-Leakage (shown in the Appendix), we
can see that the optimal strategy to integrate over x axis is distinct for different datasets. In Marmousi, using kernel $\sin(n\pi t)\cos(n\pi x)$ can improve the performance a lot. This kernel, however, does not perform well on other datasets (e.g., Kimberlina-Leakage).  

%Overall, our method is not only supported by theory, but also supported by experimental results. %Moreover, the experiment about $\sin(n\pi t)$ kernel proves the necessity of high frequency components. In Kimberlina Leakage, using a 2D sin kernel $\sin(n_S\pi t)+\sin(n_S\pi x)$ have performance improvement. However, such an improvement is limited, and does not work well in all datasets. only transforming time axis into frequency domain
\begin{table}[h]
\scriptsize
\setlength{\tabcolsep}{1.2mm}
\begin{tabular}{l|l|l|l|l}
\hline
{\textbf{Dataset}} & {\textbf{Encoder Kernel}} & {\textbf{MAE$\downarrow$}} & {\textbf{MSE$\downarrow$}} & {\textbf{SSIM$\uparrow$}}   \\ \hline
\multirow{6}{*}{Marmousi}                                                      & $\sin(n\pi t)\mathbbm{1}(x)/(x_{max}-x_{min})$                 & 136.67   & 36003.43 & 0.7972 \\ \cline{2-5}
                                                                               & $\sin(n\pi t)\sin(n\pi x)$  & 138.76 & 37648.80 & 0.8042 \\ \cline{2-5}
                                                                               & $\sin(n\pi t)\cos(n\pi x)$  & \textbf{128.33} & \textbf{32451.22} & \textbf{0.8115} \\ \cline{2-5}
                                                                               & $\cos(n\pi t)\sin(n\pi x)$  & 140.14 & 38417.23 & 0.8031 \\ \cline{2-5}
                                                                               & $\sin(n\pi(x+t))$             & 141.58 & 38383.58 & 0.7892 \\ \cline{2-5} 
                                                                               & $\sin(n\pi t)+\sin(n\pi x)$ & 142.12 & 38261.56 & 0.7884 \\ \hline 
\end{tabular}
\caption{Quantitative results for Different Encoder Kernel.}
\label{enkernel}
\end{table}
% & $\sin(n\pi t)$               & 155.38 & 44928.37 & 0.7803 \\ \cline{2-5}

\textbf{Different Decoder Kernels}

Then, we test different kernels for geophysical properties. In particular, we evaluate a series of 2D kernels: different 2D sine kernels, a sinc function kernel ($\sin(\pi\|\mathbf{r}-\mu_m\|_2)/\|\mathbf{r}-\mu_m\|_2$), and a Gaussian kernel with a smaller variance, noted as Gaussian$\sigma$. For the sinc function, the choice of $\mu_m$ is the same as the Gaussian kernel, while for Gaussian$\sigma$, we choose $\sigma$ as 1/3 of the original. The quantitative results are shown in Table~\ref{dekernel}. As we can see, our choice of kernel outperforms the rest kernels. A smaller variance of Gaussian will yield a slightly worse result, while sinc kernel performs similarly to the Gaussian$\sigma$.
\begin{table}[h]
\scriptsize
\setlength{\tabcolsep}{1.8mm}
\begin{tabular}{l|l|l|l|l}
\hline
{\textbf{Dataset}} & {\textbf{Decoder Kernel}} & {\textbf{MAE$\downarrow$}} & {\textbf{MSE$\downarrow$}} & {\textbf{SSIM$\uparrow$}} \\ \hline
\multirow{6}{*}{Marmousi}                                                       & Gaussian                        & \textbf{136.67}   & \textbf{36003.43} & \textbf{0.7972} \\ \cline{2-5}
                                                                                & Sinc                            & 138.02 & 36534.44 & 0.7952 \\ \cline{2-5} 
                                                                                & Gaussian$\sigma$                & 138.19 & 36579.46 & 0.7954 \\ \cline{2-5}
                                                                                & $\sin(n\pi x)\sin(n\pi z)$  & 177.36 & 56102.75 & 0.7455 \\ \cline{2-5}
                                                                                & $\cos(n\pi x)\sin(n\pi z)$  & 165.38 & 49463.79 & 0.7491 \\ \cline{2-5}
                                                                                & $\sin(n\pi x)\cos(n\pi z)$  & 175.92 & 55424.26 & 0.7376 \\ \cline{2-5}
                                                                                & $\sin(n\pi(x+z))$             & 209.47 & 74167.16 & 0.7057 \\ \cline{2-5} 
                                                                                & $\sin(n\pi x)+\sin(n\pi z)$ & 216.12 & 78496.77 & 0.7030 \\ \hline 
\end{tabular}
\caption{Quantitative results for Different Decoder Kernel.}
\label{dekernel}
\end{table}

\textbf{Different Number of Kernels}

\begin{figure}[!ht]
\begin{center}
\vspace{-1mm}
\centerline{\includegraphics[width=8cm]{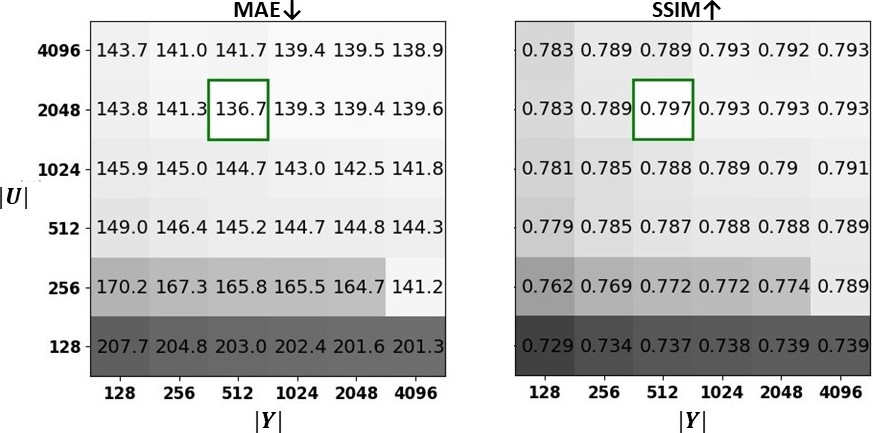}}
\vskip -0.1in
\caption{Performance over dimensions of $\bm{U}$ and $\bm{Y}$.} 
\label{grid}
\end{center}
\vskip -0.2in
\end{figure}

We also test different numbers of kernels for both Sine and Gaussian. We evaluate performance over a 6$\times$6 grid where the dimensions of $\bm{U}$ and $\bm{Y}$ vary from 128 to 4096. The quantitative results are shown in Figure~\ref{grid}. Results indicate that the current selection of dimensions is appropriate. Obviously, reducing the model's size reduces its capacity, while choices of higher dimension are more prone to overfit. However, choosing a small dimension yields a smaller number of parameters and FLOPs. One can easily balance the performance and the cost based on his requirements and resources, indicating the flexibility of our model.

% \begin{table}[h]
% \scriptsize
% \setlength{\tabcolsep}{2.9mm}
% \begin{tabular}{l|l|l|l|l}
% \hline
% {\textbf{Dataset}} & {\textbf{\#kernel}} & {\textbf{MAE$\downarrow$}} & {\textbf{MSE$\downarrow$}} & {\textbf{SSIM$\uparrow$}} \\ \hline
% \multirow{5}{*}{Marmousi}           & N=2048; M=512   & \textbf{136.67}   & \textbf{36003.43} & \textbf{0.7972} \\ \cline{2-5}
%                                     & N=1024; M=512   & 144.74 & 39568.40 & 0.7876 \\ \cline{2-5} 
%                                     & N=4096; M=512   & 141.69 & 38057.64 & 0.7890 \\ \cline{2-5} 
%                                     & N=2048; M=128   & 143.80 & 38900.92 & 0.7832 \\ \cline{2-5} 
%                                     & N=2048; M=1024  & 139.28 & 37179.20 & 0.7929 \\ \hline
% \end{tabular}
% \caption{Quantitative results for Different Number of Kernels.}
% \label{num_kernel}
% \end{table}

\textbf{Different Decoder Architectures}

We aim to design an effective and efficient decoder to reverse the integral transform over a velocity map. The shifted Gaussian kernels used in integral transform split the velocity map into overlapping windows and encode the local structure within each window. To reconstruct the global structure of the velocity map from these local features, we leverage the transformer's power in modeling long-range interaction in \textit{a single layer}. Options like conv/deconv require more layers to cover long range.

To better illustrate this, we test the performance of different decoder architecture. Results are provided in Table~\ref{arch}. A transformer layer followed by a linear layer is a more accurate decoder than shallow conv/deconv layers. Deeper decoders with more conv/deconv layers achieve more accurate results, but require a larger model. When using the deconv decoder of InversionNet in our method, we achieve better performance, clearly outperforming InversionNet (MAE 126.6 vs. 149.7). %Results in Kimberlina-Leakage show a same pattern.

%: (1) replacing the transformer with a two-layers CNN; (2) using a large deconvolution layer as the decoder; (3) using three-layer interpolations plus convolution layers as the decoder. The quantitative results are shown in Table~\ref{arch}. The results demonstrate that although our method is model-agnostic in a certain degree (e.g., one can replace transformer with other networks), the current decoder architecture is necessary.
\begin{table}[h]
\scriptsize
\setlength{\tabcolsep}{0.3mm}
\begin{tabular}{l|l|r|r|r|r|r}
\hline
{\textbf{Dataset}} & {\textbf{Architecture}} & {\textbf{MAE$\downarrow$}} & {\textbf{MSE$\downarrow$}} & {\textbf{SSIM$\uparrow$}}  & {\textbf{\#Params}} & {\textbf{FLOPs}} \\ \hline
\multirow{4}{*}{Marmousi}           & Transformer$\times 1$ + Linear*   & 136.67   & 36003.43 & 0.7972 & 1.45M         & 9.3M\\ \cline{2-7}
                                    & Conv $\times 2$ + Linear          & 140.72 & 37345.58  & 0.7903  & 0.30M      & 9.2M\\ \cline{2-7}
                                    & Deconv $\times 1$ + Linear        & 167.98 & 49728.14  & 0.7520  & 0.35M       & 10.8M\\ \cline{2-7}
                                    & (Deconv + Conv) $\times 5$        & \textbf{126.59}   & \textbf{33830.73} & \textbf{0.8158}  & 12.71M        & 94.6M   \\ \cline{2-7}
                                    %& (Up + Conv) $\times 1$            & 400.47 & 304996.66 & 0.5908 \\ \cline{2-7}
                                    & (Up + Conv) $\times 5$            & 128.74   & 34854.78 & 0.8120  & 4.01M        & 56.7M \\ \hline
\end{tabular}
\caption{Comparison among different decoder structures. (*) indicates the default decoder option.}
\label{arch}
\end{table}

\textbf{Results for a Larger Decoder}

Here, we evaluate our method with a larger/deeper decoder. Firstly, we test it using multiple unshared linear layers, rather than a shared one $L_{r1}$, in the last part of our decoder. Furthermore, we evaluate our model with a deeper transformer. The quantitative results are shown in Table~\ref{larger}. The result using unshared linear layers indicates that a single linear layer is enough and the model does not benefit from more parameters. On the other hand, a deeper transformer can improve the performance. Similar to the number of kernels, the balance is based on requirements.

\begin{table}[h]
\scriptsize
\setlength{\tabcolsep}{2.6mm}
\begin{tabular}{l|l|l|l|l}
\hline
{\textbf{Dataset}} & {\textbf{Architecture}} & {\textbf{MAE$\downarrow$}} & {\textbf{MSE$\downarrow$}} & {\textbf{SSIM$\uparrow$}} \\ \hline
\multirow{4}{*}{Marmousi}           & 1 layer Transformer  & 136.67   & 36003.43 & 0.7972 \\ \cline{2-5}
                                    & Multi-Linear         & 138.82 & 36801.89 & 0.7939 \\ \cline{2-5} 
                                    & 2 layers Transformer & 134.24 & 35111.23 & 0.8002 \\ \cline{2-5} 
                                    & 3 layers Transformer & \textbf{132.19} & \textbf{34502.25} & \textbf{0.8037} \\ \hline
\end{tabular}
\caption{Quantitative results for a Larger Decoder.}
\label{larger}
\end{table}

\subsection{Singular Value Analysis}
\begin{figure}[!ht]
\begin{center}
\centerline{\includegraphics[width=\columnwidth]{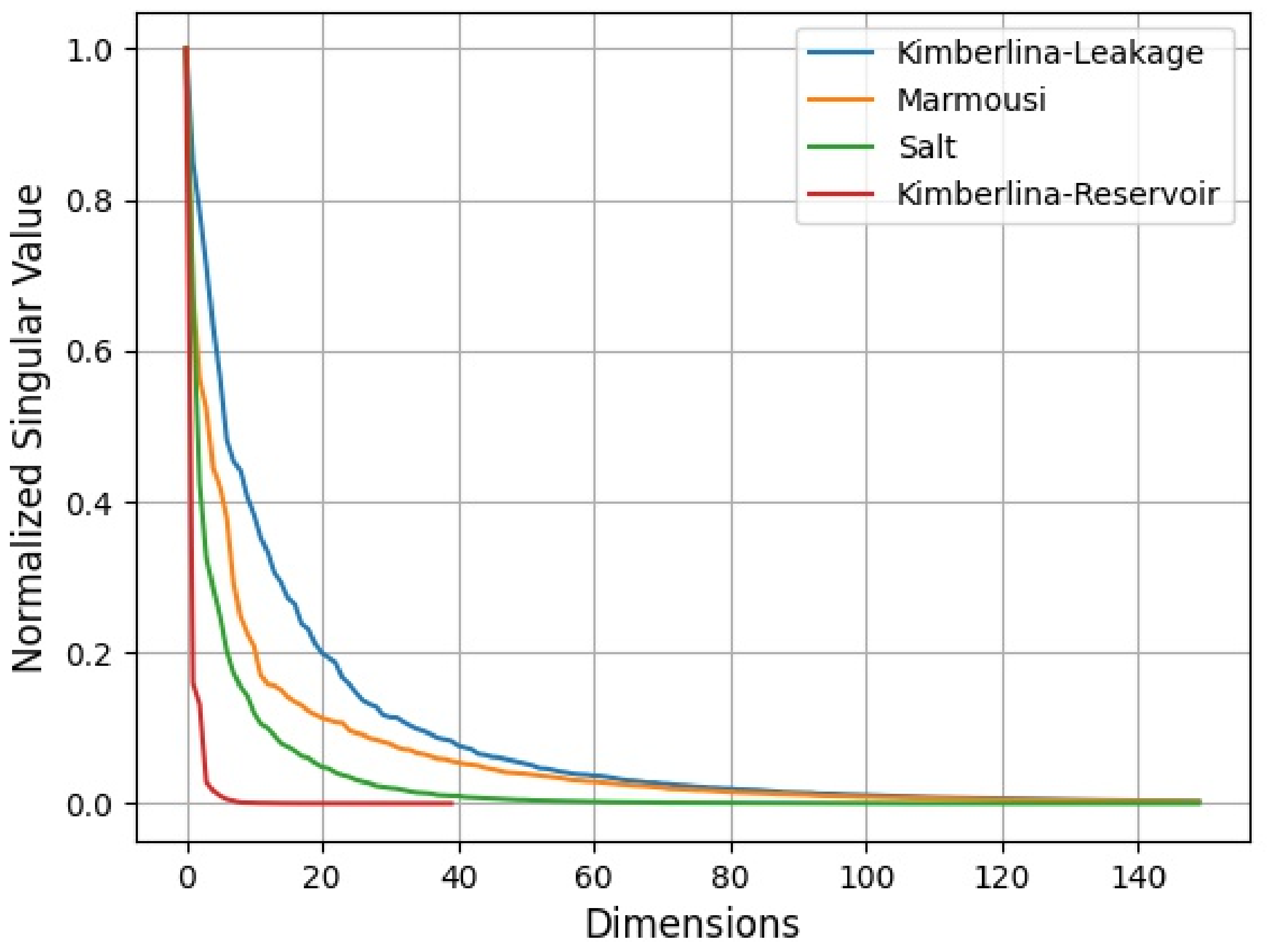}}
\caption{Normalized Singular Value Decomposition of the linear layer on different datasets.} 
\label{svd}
\end{center}
\vskip -0.2in
\end{figure}

Another major benefit of our simplified model is the ease of analysis. Since we use only one linear layer in the encoder, we can analyze it by performing singular value decomposition. The results are shown in Figure~\ref{svd}. Since the singular value varies greatly in different datasets, we divide it by its maximum value to normalize it and trunk it at 150 dim. Results indicate that for all datasets, the number of essential dimensions is less than 100. In other words, a 100-dimensional latent space is sufficient to represent the data. Specifically, we can see that a ten-dimensional latent space is enough for Kimberlina-Reservoir dataset. That answers why the required number of parameters of both our InvLINT and InversionNet on Kimberlina-Reservoir datasets are much smaller than that on other datasets. All in all, with such a simple architecture, our InvLINT is able to not only help in analyzing the problem but also help us quantify the difficulty of different datasets.

% \subsection{Robustness}
% In this part, we validate the robustness of our InvLINT models by two additional experiments: (1) testing data contaminated by Gaussian noise and (2) testing data with missing traces. The quantitative results are shown in Appendix. The result demonstrates that, at the cost of simplifying the model, our model is equally sensitive to noise 

\subsection{Comparison to traditional FWI:}
We performed new comparison with a widely used traditional FWI method (i.e., Multiscale FWI \cite{virieux2009overview}) on three seismic FWI datasets (Marmousi, Kimberlina-Leakage, Salt). 
Our method is consistently better on 3 datasets (MAE: 11.7 vs. 42.0 in Kimberlina-Leakage, 140.7 vs. 199.5 in Marmousi, 26.1 vs. 176.6 in Salt). The traditional FWI requires a good initial guess and optimization per sample, resulting in slow processing (e.g., 4 hours per sample in Kimberlina-Leakage). Due to the limited rebuttal duration, we ran the comparison over 5 samples per dataset.

\section{Related works}
%\subsection{Physics-driven Methods}
\subsection{Data-driven Methods for FWI}
Recently, based on deep learning, a new type of method has been developed. \citet{araya2018deep} use a fully connected network to invert velocity maps in FWI. \citet{wu2019inversionnet} consider the FWI as an image-to-image translation problem, and employ encoder-decoder CNN to solve. By using generative adversarial networks (GANs) and transfer learning, \citet{zhang2019velocitygan} achieved improved performance. In \citet{zeng2021inversionnet3d}, authors present an efficient and scalable encoder-decoder network for 3D FWI. \citet{feng2021multiscale} develop a multi-scale framework with two convolutional neural networks to reconstruct the low- and high-frequency components of velocity maps. A thorough review on deep learning for FWI can be found in \citet{adler2021deep}.

\subsection{Physics-informed machine learning}
Previous pure data-driven methods can be considered as incorporating physic information in training data. On the other hand, integrating the physic knowledge into loss function or network architecture is another direction. All of them are called Physics-informed neural networks~(PINN). Raissi et al. proposed utilizing nonlinear PDEs in the loss function as a soft constrain \cite{raissi2019physics}. Through a hard constraint projection, Chen et al. proposed a framework to ensure model's predictions strictly conform to physical mechanisms \cite{chen2021theory}. Based on the universal approximation theorem of operators, in \citet{lu2021learning}, authors proposed DeepONet to learn continuous operators or complex systems. \citet{sun2021physics} proposed a hybrid network design, which involves deterministic, physics-based modeling and data-driven deep learning. A comprehensive review of PINN can be found in \citet{karniadakis2021physics}.

%\section{Discussion}

\section{Conclusion}
In this paper, we find an intriguing property of geophysics inversion: \textit{a near-linear relationship between the input and output, after applying integral transform in high dimensional space.} Furthermore, this property can be easily turned into a light-weight encoder-decoder network for inversion. The encoder contains the integration of seismic data and the linear transformation without fine-tuning. The decoder consists of a single transformer block to reverse the integral of velocity with Gaussian kernels.

Experiments show that this interesting property holds for two geophysics inversion problems over four different datasets. Compared to much deeper InversionNet, our method achieves comparable accuracy, but consumes significantly fewer parameters.

\bibliography{main}
\bibliographystyle{icml2021}

\newpage
\appendix

\section{Appendix}
\subsection{Inversion Results of Different Datasets}
\begin{figure}[!ht]
\begin{center}
\centerline{\includegraphics[width=\columnwidth]{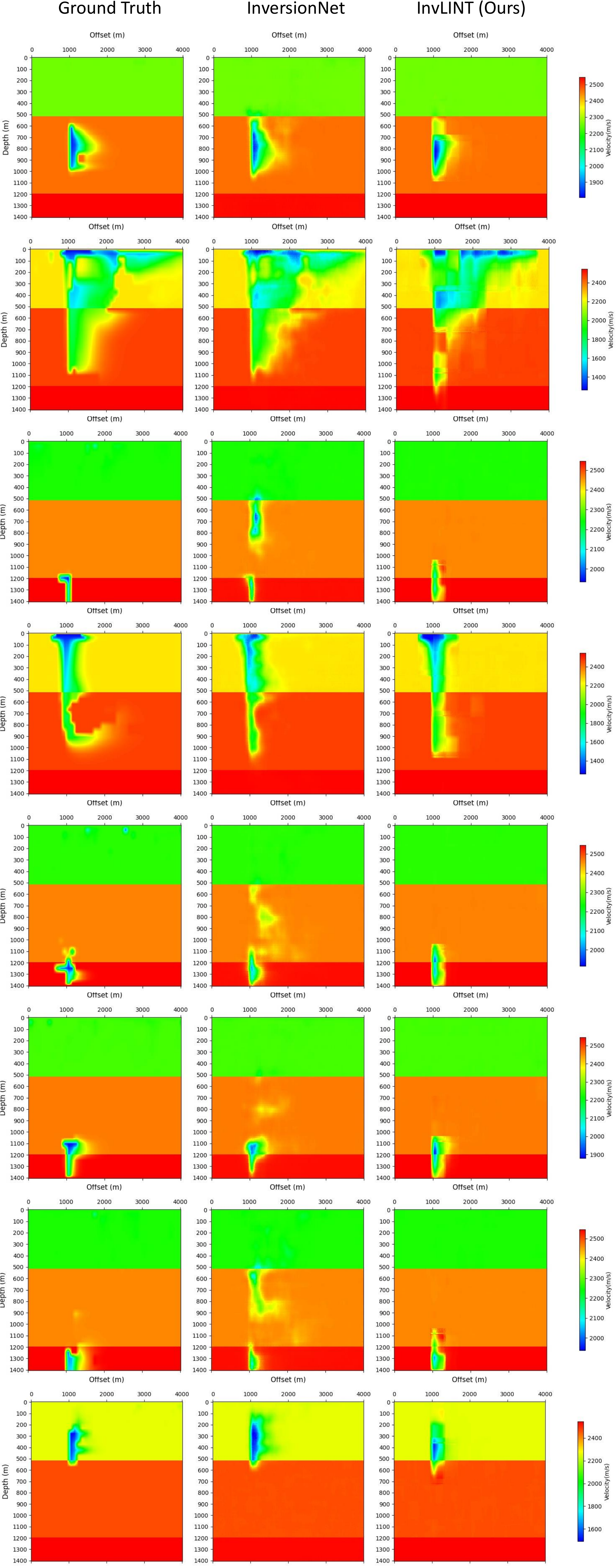}}
\vskip -0.12in
\caption{Illustration of results evaluated on Kimberlina Leakage} %(a) Results on Kimerlina CO$_2$; (b) Results on Marmousi; (c) Results on Salt; (d) Results on Kimberlina-Reservoir. Our InvLINT achieves good results on all four datasets.
\vskip -0.2in
\end{center}
\vskip -0.2in
\end{figure}

\begin{figure}[!ht]
\begin{center}
\centerline{\includegraphics[width=\columnwidth]{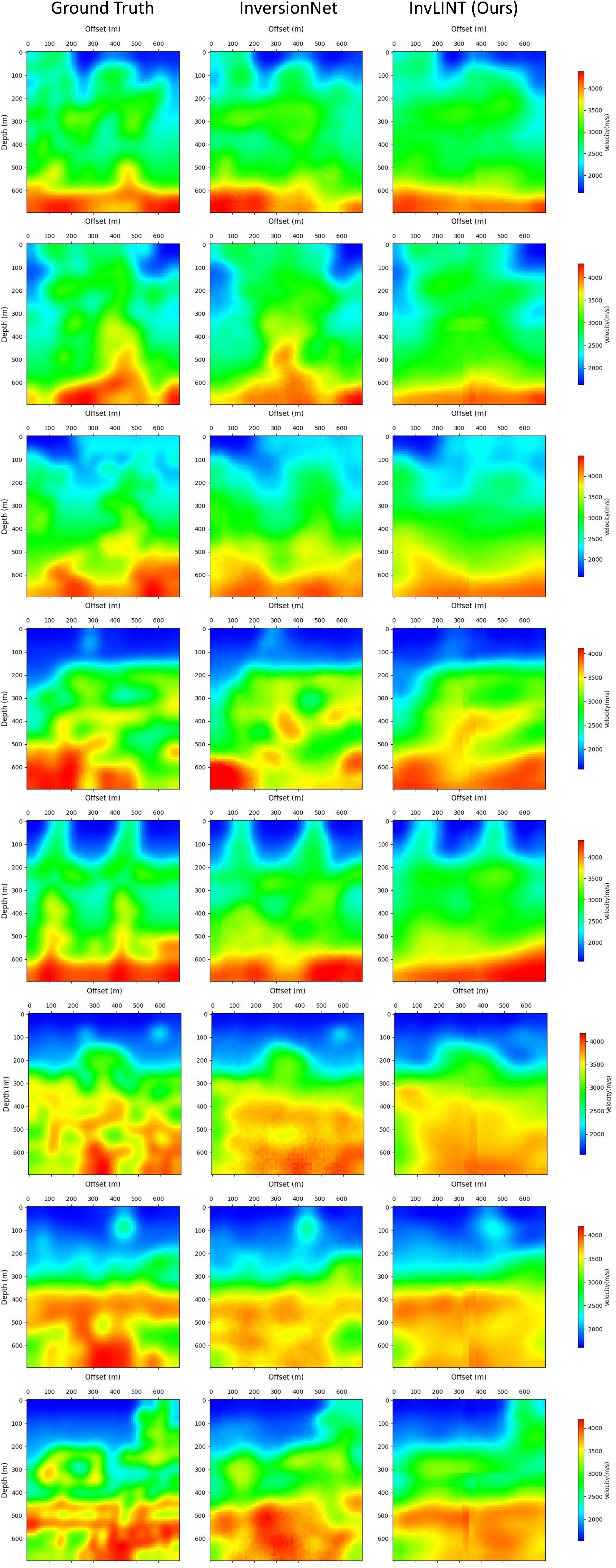}}
\caption{Illustration of results evaluated on Marmousi} %(a) Results on Kimerlina CO$_2$; (b) Results on Marmousi; (c) Results on Salt; (d) Results on Kimberlina-Reservoir. Our InvLINT achieves good results on all four datasets.
\end{center}
\vskip -0.2in
\end{figure}

\begin{figure}[!ht]
\begin{center}
\centerline{\includegraphics[width=\columnwidth]{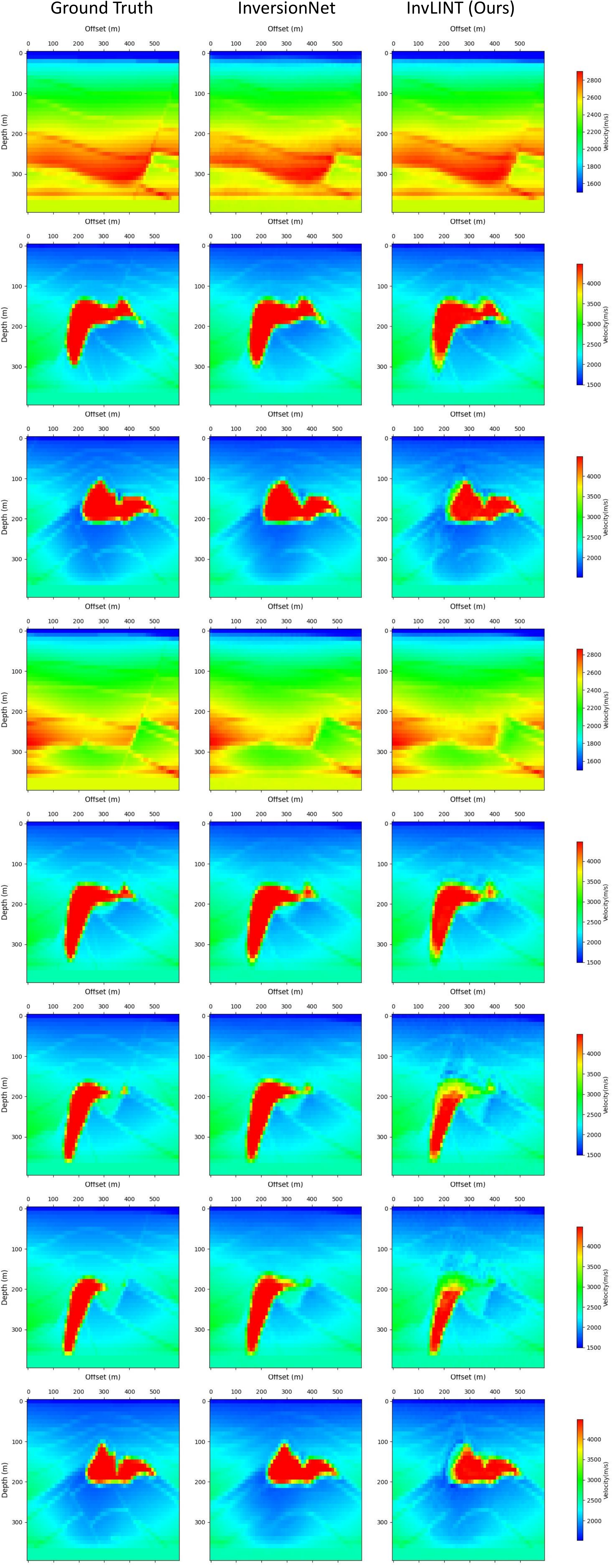}}
\caption{Illustration of results evaluated on Salt} %(a) Results on Kimerlina CO$_2$; (b) Results on Marmousi; (c) Results on Salt; (d) Results on Kimberlina-Reservoir. Our InvLINT achieves good results on all four datasets.
\end{center}
\vskip -0.2in
\end{figure}

\begin{figure}[!ht]
\begin{center}
\centerline{\includegraphics[width=\columnwidth]{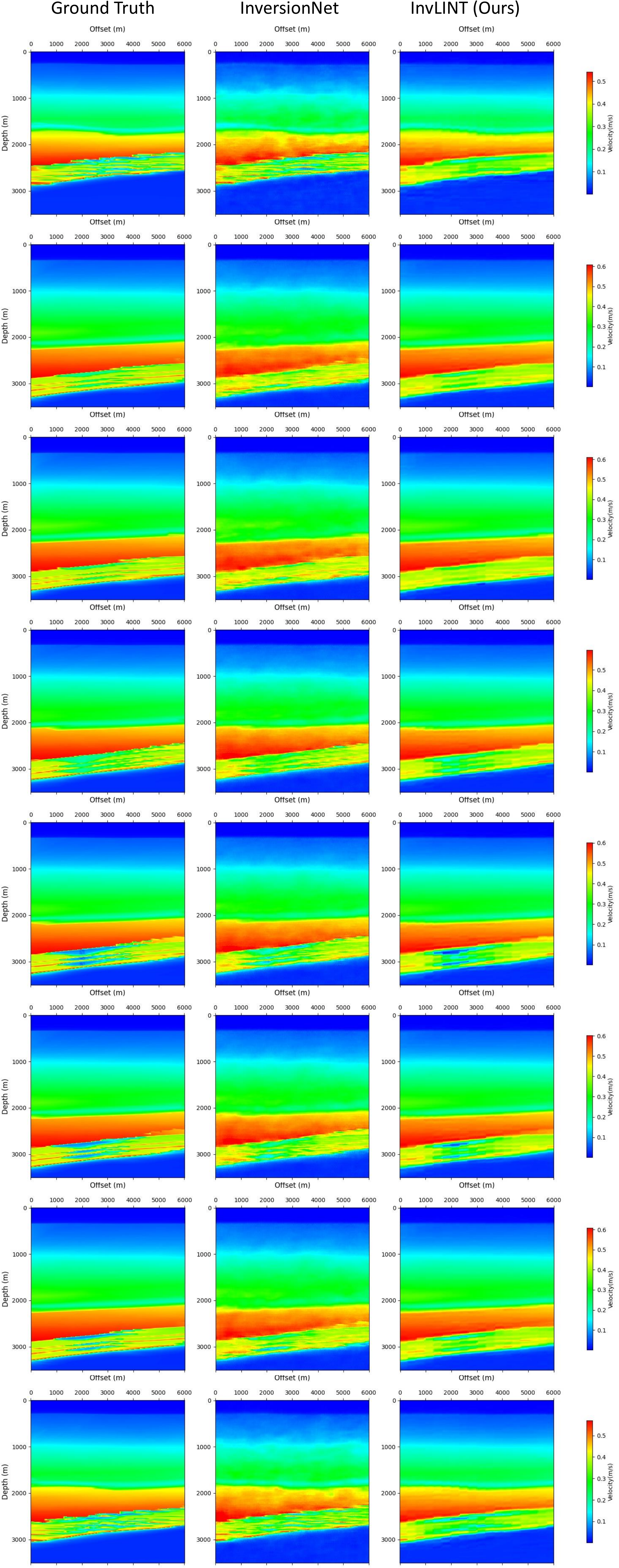}}
\caption{Illustration of results evaluated on Kimberlina Reservoir} %(a) Results on Kimerlina CO$_2$; (b) Results on Marmousi; (c) Results on Salt; (d) Results on Kimberlina-Reservoir. Our InvLINT achieves good results on all four datasets.
\end{center}
\vskip -0.2in
\end{figure}

\subsection{Ablation Test on Kimberlina Leakage}
The ablation test results on Kimberlina Leakage are shown in Table \ref{app1}-\ref{app5}
\begin{table}[!h]
\scriptsize
\setlength{\tabcolsep}{1.2mm}
\begin{tabular}{l|l|l|l|l}
\hline
{\textbf{Dataset}} & {\textbf{Encoder Kernel}} & {\textbf{MAE$\downarrow$}} & {\textbf{MSE$\downarrow$}} & {\textbf{SSIM$\uparrow$}}   \\ \hline
\multirow{6}{*}{\begin{tabular}[c]{@{}l@{}}Kimberlina \\ Leakage\end{tabular}} & $\sin(n\pi t)\mathbbm{1}(x)/(x_{max}-x_{min})$                 & 8.13     & \textbf{1534.60}           & 0.9812 \\ \cline{2-5}
                                                                               & $\sin(n\pi t)\sin(n\pi x)$  & 11.07  & 3227.71  & 0.9783 \\ \cline{2-5}
                                                                               & $\sin(n\pi t)\cos(n\pi x)$  & 8.88  & 2015.23  & 0.9804 \\ \cline{2-5}
                                                                               & $\cos(n\pi t)\sin(n\pi x)$  & 10.95  & 3222.21  & 0.9782 \\ \cline{2-5}
                                                                               & $\sin(n\pi(x+t))$             & 8.17   & 1751.89  & 0.9815 \\ \cline{2-5} 
                                                                               & $\sin(n\pi t)+\sin(n\pi x)$ & \textbf{8.10}   & 1760.43  & \textbf{0.9817} \\ \hline
\end{tabular}
\caption{Quantitative results for Different Encoder Kernel.}
\label{app1}
\end{table}
%& $\sin(n\pi t)$               & 9.41   & 2264.24  & 0.9802 \\ \cline{2-5}

\begin{table}[!h]
\scriptsize
\setlength{\tabcolsep}{2.2mm}
\begin{tabular}{l|l|l|l|l}
\hline
{\textbf{Dataset}} & {\textbf{Decoder Kernel}} & {\textbf{MAE$\downarrow$}} & {\textbf{MSE$\downarrow$}} & {\textbf{SSIM$\uparrow$}} \\ \hline
\multirow{6}{*}{\begin{tabular}[c]{@{}l@{}}Kimberlina \\ Leakage\end{tabular}}  & Gaussian                        & \textbf{8.13}     & \textbf{1534.60}           & \textbf{0.9812} \\ \cline{2-5}
                                                                                & Sinc                            & 8.90   & 2051.99  & 0.9789 \\ \cline{2-5} 
                                                                                & Gaussian$\sigma$                & 8.84   & 2042.94  & 0.9790 \\ \cline{2-5}
                                                                                & $\sin(n\pi x)\sin(n\pi z)$  & 15.41  & 7357.48  & 0.9764 \\ \cline{2-5}
                                                                                & $\cos(n\pi x)\sin(n\pi z)$  & 15.40  & 7349.02  & 0.9764 \\ \cline{2-5}
                                                                                & $\sin(n\pi x)\cos(n\pi z)$  & 15.37  & 7252.45  & 0.9765 \\ \cline{2-5}
                                                                                & $\sin(n\pi(x+z))$             & 12.86  & 4721.48  & 0.9767 \\ \cline{2-5} 
                                                                                & $\sin(n\pi x)+\sin(n\pi z)$ & 13.21  & 74719.95 & 0.9764 \\ \hline
\end{tabular}
\caption{Quantitative results for Different Decoder Kernel.}
\label{app2}
\end{table}

\begin{table}[!h]
\scriptsize
\setlength{\tabcolsep}{3.1mm}
\begin{tabular}{l|l|l|l|l}
\hline
{\textbf{Dataset}} & {\textbf{\#kernel}} & {\textbf{MAE$\downarrow$}} & {\textbf{MSE$\downarrow$}} & {\textbf{SSIM$\uparrow$}} \\ \hline
\multirow{5}{*}{\begin{tabular}[c]{@{}l@{}}Kimberlina \\ Leakage\end{tabular}} & N=2048; M=512  & \textbf{8.13}     & \textbf{1534.60}      & \textbf{0.9812} \\ \cline{2-5}
                                    & N=1024; M=512   & 8.63   & 1946.62  & 0.9811 \\ \cline{2-5} 
                                    & N=4096; M=512   & 8.29   & 1780.75  & 0.9808 \\ \cline{2-5} 
                                    & N=2048; M=128   & 8.76   & 2007.14  & 0.9805 \\ \cline{2-5} 
                                    & N=2048; M=1024  & 8.59   & 1898.95  & 0.9808 \\ \hline
\end{tabular}
\caption{Quantitative results for Different Number of Kernels.}
\label{app3}
\end{table}

\begin{table}[!h]
\scriptsize
\setlength{\tabcolsep}{2.2mm}
\begin{tabular}{l|l|l|l|l}
\hline
{\textbf{Dataset}} & {\textbf{Architecture}} & {\textbf{MAE$\downarrow$}} & {\textbf{MSE$\downarrow$}} & {\textbf{SSIM$\uparrow$}} \\ \hline
\multirow{4}{*}{\begin{tabular}[c]{@{}l@{}}Kimberlina \\ Leakage\end{tabular}} & Transformer$\times 1$ + Linear*  & 8.13     & 1534.60      & 0.9812 \\ \cline{2-5}
                                    & Conv $\times 2$ + Linear                 & 13.42  & 2447.81   & 0.9762 \\ \cline{2-5}
                                    & Deconv $\times 1$ + Linear               & 21.32  & 4919.03   & 0.9648 \\ \cline{2-5} 
                                    & (Deconv + Conv) $\times 5$               & \textbf{6.86}  & \textbf{1462.29}   & \textbf{0.9841} \\ \cline{2-5}
                                    & (Up + Conv) $\times 5$                   & 6.87 & 1516.80  & 0.9840 \\ \hline
                                    %& Interpolation + Conv & 84.96  & 21841.89  & 0.8951 \\ \hline
\end{tabular}
\caption{Quantitative results for Different Decoder Architectures. (*) indicates the default decoder option.}
\label{app4}
\end{table}

\begin{table}[!h]
\scriptsize
\setlength{\tabcolsep}{2.7mm}
\begin{tabular}{l|l|l|l|l}
\hline
{\textbf{Dataset}} & {\textbf{Architecture}} & {\textbf{MAE$\downarrow$}} & {\textbf{MSE$\downarrow$}} & {\textbf{SSIM$\uparrow$}} \\ \hline
\multirow{4}{*}{\begin{tabular}[c]{@{}l@{}}Kimberlina \\ Leakage\end{tabular}} & 1 layer Transformer  & 8.13     & \textbf{1534.60}      & 0.9812 \\ \cline{2-5}
                                    & Multi-Linear         & 8.14   & 1799.70  & 0.9811 \\ \cline{2-5} 
                                    & 2 layers Transformer & 8.24   & 1781.50  & 0.9812 \\ \cline{2-5} 
                                    & 3 layers Transformer & \textbf{8.09}   & 1707.53  & \textbf{0.9813} \\ \hline
\end{tabular}
\caption{Quantitative results for a Larger Decoder.}
\label{app5}
\end{table}

\subsection{Regression Results for the Encoder Linear Layer}
We also show here the regression results of the linear layer in our encoder on different datasets in Table \ref{regression}. As a reference, we also show the range and mean of the regression target value as $y_{max}$-$y_{min}$,  $|y_{mean}|$. The result demonstrate that how well the regressions are fitted.
\begin{table}[ht]
\scriptsize
\setlength{\tabcolsep}{1.2mm}
\begin{tabular}{l|l|l|l|l|l}
\hline
{\textbf{Dataset}} & {\textbf{Set}} & {\textbf{MAE$\downarrow$}} & {\textbf{MSE$\downarrow$}} & {\textbf{$y_{max}-y_{min}$}} & $|y_{mean}|$ \\ \hline
\multirow{2}{*}{\begin{tabular}[c]{@{}l@{}}Kimberlina \\ Leakage\end{tabular}} & Training set & 2.83   & 45.48     & 884.4           & 490.93    \\ \cline{2-6} 
                                & Test set     & 4.63   & 245.09    & 885.27          & 492.65    \\ \hline
\multirow{2}{*}{Marmousi}       & Training set & 4.26   & 33.38     & 107.05          & 4.92      \\ \cline{2-6} 
                                & Test set     & 4.29   & 33.9      & 103.01          & 5.1       \\ \hline
\multirow{2}{*}{Salt}           & Training set & 0.28   & 0.46      & 51.3            & 12.11     \\ \cline{2-6} 
                                & Test set     & 0.48   & 1.98      & 49.35           & 11.97     \\ \hline
\multirow{2}{*}{\begin{tabular}[c]{@{}l@{}}Kimberlina \\ Reservoir\end{tabular}}             & Training set & 169.25 & 2607.27   & 26497.1         & 7197.4873 \\ \cline{2-6} 
                                & Test set     & 212.64 & 109288.08 & 26496.956       & 6849.95   \\ \hline
\end{tabular}
\caption{Quantitative results for a Larger Decoder.}
\label{regression}
\end{table}

\end{document}